\documentclass[letterpaper, 10 pt, conference]{ieeeconf}  
\IEEEoverridecommandlockouts                                    

\usepackage{bm}

\RequirePackage[loading]{tracefnt}

\usepackage{graphicx}                       
\usepackage{graphics}                       
\usepackage{epsfig}                         
\usepackage[tight,footnotesize]{subfigure}  
\graphicspath{./pics/RAL_fig_for_revise}
\graphicspath{./pics/RAL_new_fig}
\usepackage{amssymb,amsmath}
\usepackage{mdwmath}
\usepackage{commath}   
\usepackage{eqparbox}
\usepackage{mathtools}
\usepackage[utf8]{inputenc} 
\usepackage[english]{babel}
\usepackage{amsmath,amsfonts,amssymb}

\newcommand{\nn}{\nonumber}

\newcommand{\beq}{\begin{equation}}
\newcommand{\eeq}{\end{equation}}
\newcommand{\bear}{\begin{eqnarray}}
\newcommand{\bears}{\begin{eqnarray*}}
\newcommand{\eear}{\end{eqnarray}}
\newcommand{\eears}{\end{eqnarray*}}
\newcommand{\bdm}{\begin{displaymath}}
\newcommand{\edm}{\end{displaymath}}
\newcommand{\lba}{\left[\begin{array}}
\newcommand{\ear}{\end{array}\right]}

\usepackage{tabularray}
\UseTblrLibrary{diagbox}

\usepackage{stfloats}                       
\usepackage{url} 
\usepackage{hyperref}
\usepackage{cite}                           
\usepackage[T1]{fontenc} 
\usepackage{bbm}

\usepackage{tabularx}
\usepackage{multirow}
\usepackage[table]{xcolor}
\usepackage{longtable}
\usepackage{booktabs} 			
\usepackage{xcolor,colortbl}    

\usepackage{multicol}
\usepackage{graphicx}
\usepackage{placeins}

\usepackage[T1]{fontenc}
\usepackage{etoolbox}

\usepackage{tabularx}
\usepackage{paralist}

\usepackage{multirow}
\usepackage{makecell}

\makeatletter
\patchcmd{\@maketitle}{\normalsize}{\normalsize}{}{}
\makeatother

\usepackage{etoolbox}
\makeatletter
\patchcmd{\@makecaption}
  {\scshape}
  {}
  {}
  {}
\makeatother
\title{\fontsize{24}{29} \textrm{Attention for Robot Touch: Tactile Saliency\\ Prediction for Robust Sim-to-Real Tactile Control} }
\author{
Yijiong Lin, 
Mauro Comi,
Alex Church,
Dandan Zhang,
Nathan F. Lepora \\
\thanks{
YL was supported by a UoB-CSC-joint scholarship. NL was supported by a Leadership Award from the Leverhulme Trust on ‘A biomimetic forebrain for robot touch’ (RL-2016-39).
{\em (Corresponding author: Yijiong Lin)}
}
\thanks{All authors are with the Department of Engineering Mathematics and Bristol Robotics Laboratory, University of Bristol, Bristol BS8 1UB, U.K. (email: \{yijiong.lin, n.lepora\}@bristol.ac.uk)}
}

\begin{document}
\maketitle
\begin{abstract}
High-resolution tactile sensing can provide accurate information about local contact in contact-rich robotic tasks. However, the deployment of such tasks in unstructured environments remains under-investigated. To improve the robustness of tactile robot control in unstructured environments, we propose and study a new concept: \textit{tactile saliency} for robot touch, inspired by the human touch attention mechanism from neuroscience and the visual saliency prediction problem from computer vision. In analogy to visual saliency, this concept involves identifying key information in tactile images captured by a tactile sensor. While visual saliency datasets are commonly annotated by humans, manually labelling tactile images is challenging due to their counterintuitive patterns. To address this challenge, we propose a novel approach comprised of three interrelated networks: 1) a Contact Depth Network (ConDepNet), which generates a contact depth map to localize deformation in a real tactile image that contains target and noise features; 2) a Tactile Saliency Network (TacSalNet), which predicts a tactile saliency map to describe the target areas for an input contact depth map; 3) and a Tactile Noise Generator (TacNGen), which generates noise features to train the TacSalNet. Experimental results in contact pose estimation and edge-following in the presence of distractors showcase the accurate prediction of target features from real tactile images. Overall, our tactile saliency prediction approach gives robust sim-to-real tactile control in environments with unknown distractors. Project page: \url{https://sites.google.com/view/tactile-saliency/}.


\end{abstract}


\section{INTRODUCTION}\label{sec:Intro}

\begin{figure}
  \centering
    \includegraphics[width=0.9\linewidth]{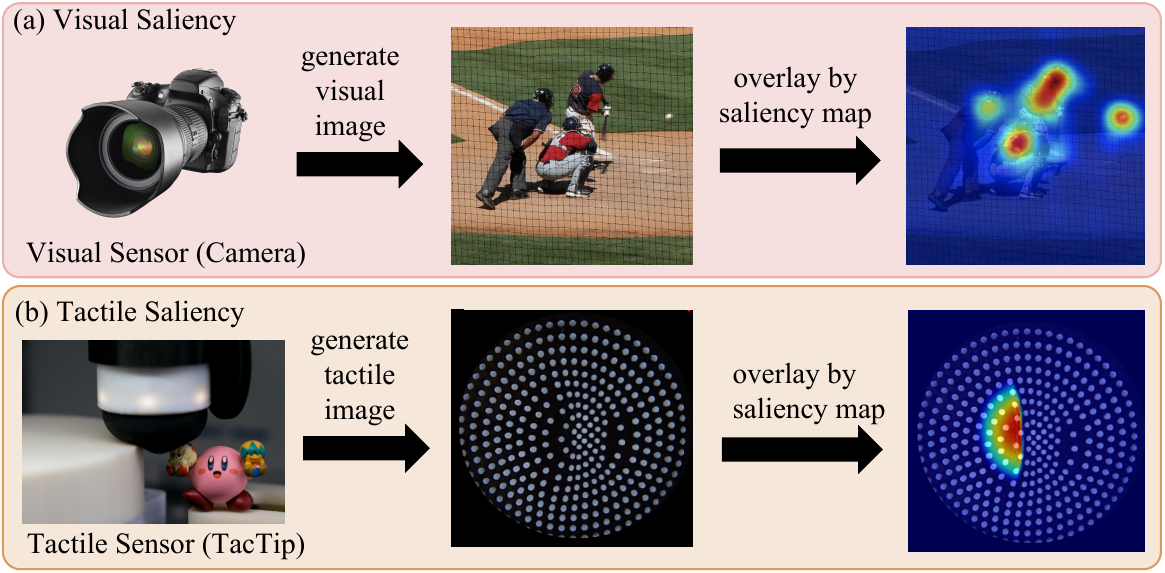}
    \vspace{-0.5em}
    \caption{Visual saliency vs tactile saliency: (a) an example of visual saliency map (right) and its source visual image (mid) from SALICON dataset \cite{jiang2015salicon}; (b) an example of tactile saliency map (right) and its source tactile image (mid) obtained by a TacTip (left) contacting a target edge (white cylinder) and a distractor (pink toy) in an edge-following task in a cluttered scene.}
  \label{fig:cover_img}
  \vspace{-0.5cm}
\end{figure}

High-resolution tactile sensing is seeing greater use in robot manipulation as a complement to vision, due to its ability to reveal fine-grained details in local contact \cite{lepora2021soft}. Despite its potential, the research community has predominantly focused on tasks with idealized experimental conditions, disregarding the impact of noise and distractors \cite{church2021tactile, Pecyna2022deformable, lin2023bitouch}. This has resulted in a limited understanding of how to achieve robust tactile control in unstructured environments, where unexpected stimuli can impair the accuracy of controllers or policies relying on tactile sensing (as shown in Fig.~\ref{fig:cover_img}b), making it difficult to achieve precise control in tactile-oriented tasks such as contour following and tactile exploration \cite{huang2022soft , kappassov2022tactile, lepora2020optimal, lin2022tactilegym2}. Therefore, it is crucial to develop a methodology that can effectively distinguish between target and noise features in tactile feedback, enabling robust tactile control in unstructured environments. 

To better describe this problem, we propose a new concept for robot touch: \textit{tactile saliency}. Analogous to \textit{visual saliency} in computer vision, we define tactile saliency to describe the critical regions of interest for a robot in a tactile image obtained by a tactile sensor. For example in Fig.~\ref{fig:cover_img}b, a tactile saliency map can indicate a target feature of edge from a real tactile image during a contour-following task in an unstructured environment. However, collecting and labelling tactile saliency data presents unique challenges compared to visual saliency data. Unlike the latter, which is typically collected through human labelling using eye trackers, mouse clicks, etc. \cite{cong2018review}, it is challenging for humans to distinguish between target and noise features in a raw tactile image. Additionally, tactile sensors are often soft and delicate, making it impractical to collect a large amount of data through direct physical contact with various stimuli while avoiding damaging the sensors, particularly if the goal is to learn the joint distribution of target and noise features.

\begin{figure*}[t!]
  \centering
    \includegraphics[width=0.85\linewidth]{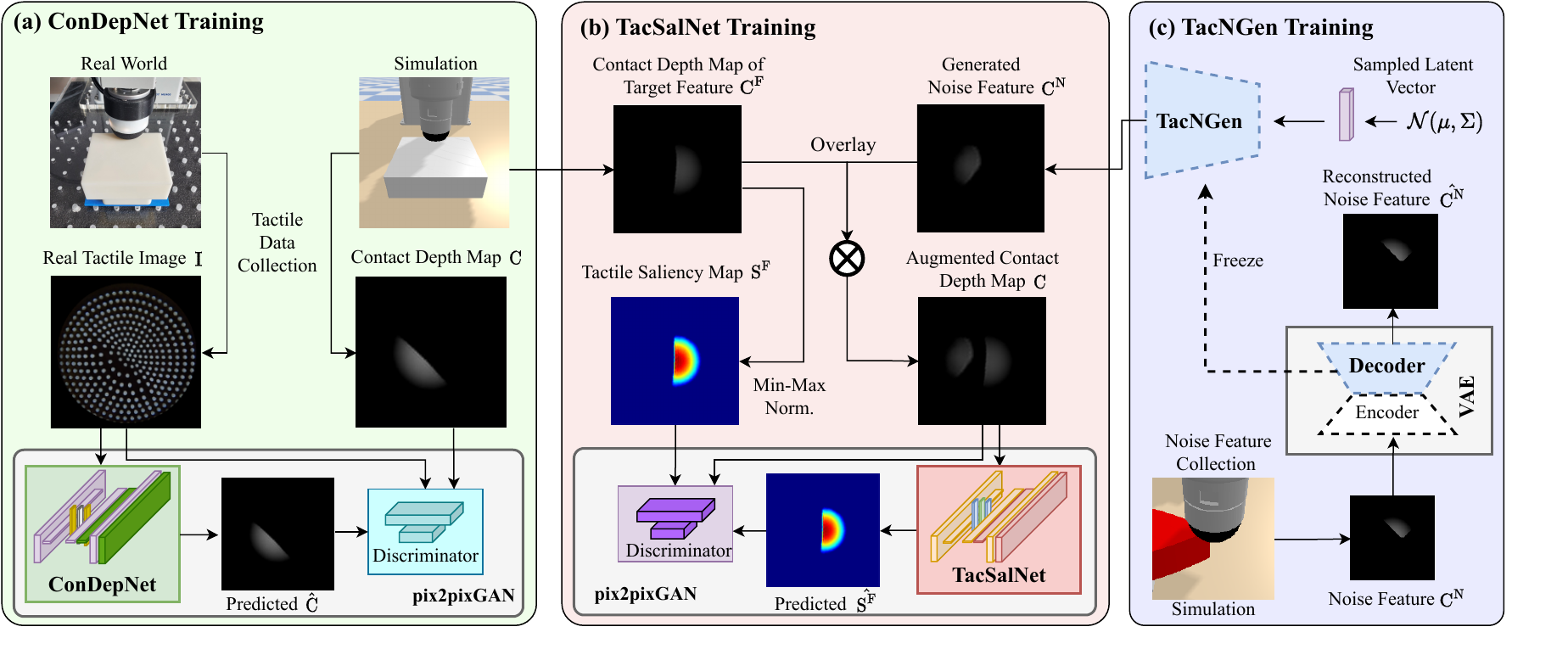}
    \vspace{-0.5em}
    \caption{Overview of the proposed 3-stage approach for tactile saliency prediction: (a) ConDepNet training for generating a contact depth map to localize deformation in a real tactile image using a dataset of paired real and simulated tactile images. (b) TacSalNet training to predict a tactile saliency map to capture a desired target feature for an input contact depth map with noise. (c) TacNGen training for generating noise features to train the TacSalNet.}
  \label{fig:method}
\vspace{-0.5em}
\end{figure*}

To address these challenges, here we propose a novel approach for tactile saliency prediction (Fig.~\ref{fig:method}) comprised of three interrelated networks: 1) a Contact Depth Network (ConDepNet), which generates a contact depth map to localize deformation for an input real tactile image that contains target and noise features in a simplified format; 2) a Tactile Saliency Network (TacSalNet), which predicts a tactile saliency map to describe the target areas for an input contact depth map; 3)~and a Tactile Noise Generator (TacNGen), which generates noise features to train the TacSalNet, making it more generalizable to unseen contact depth maps.

The proposed approach presents several advantages. Firstly, it eliminates the need for human labelling, leading to a more practical and efficient data collection process. Secondly, with the high fidelity of generated tactile noise, it enables accurate prediction of target features from real tactile images in the presence of unseen noise, enhancing the robustness of tactile control in unstructured environments. Thirdly, the TacSalNet can be seen as a simple plug-in module for various types of tactile-oriented control methods, such as pose-based tactile servoing with PID control \cite{lepora2021pose} or image-based deep reinforcement learning (deep-RL) \cite{church2021tactile, lin2023bitouch}. Moreover, this approach has the potential to apply to other types of optical tactile sensors, making it a general solution for predicting tactile saliency in a manner applicable to a wide range of uses.


The main contributions of this research are as follows:\\
1) We initiate the concept of tactile saliency for robot touch, and investigate this problem for robust sim-to-real tactile control in unstructured environments.\\
2) We propose a novel approach to predict tactile saliency from real tactile images without relying on human labelling.\\
3) We achieve precise contact pose prediction and robust sim-to-real tactile control for an edge-following task in the presence of distractors, by successfully learning a tactile saliency prediction network from simulation and applying it to the real world.
\vspace{-0.2em}

\section{RELATED WORK} \label{sec:related}




Visual saliency has been researched extensively in computer vision \cite{borji2019saliency, jia2020eml, mejjati2020look, aberman2021deep}. This concept refers to the regions or objects in an image that attracts more visual attention or are more noticeable than other regions \cite{cong2018review}. In recent years, researchers have applied this concept to touch-relevant tasks as well. Xu et al. \cite{xu2012touch} and Ni et al. \cite{ni2014touch} explored the use of visual saliency in 2D images on mobile devices to better understand and predict users' touch behaviours. Lau et al. \cite{lau2016tactile} introduced tactile mesh saliency to measure the salient parts of 3D object mesh models that are more likely to be contacted by humans. Similarly, Jiao et al. \cite{jiao2020tactile} proposed tactile sketch saliency, which uses supervised learning to jointly estimate the tactile saliency, depth map, and semantic category of a sketch. Additionally, Jain et al. \cite{Jain2021aug} utilized 3D point clouds as visual input to determine an object saliency map for touch, which guides the robot to collect more informative tactile data. Cao et al. \cite{cao2020spatio} proposed the Spatio-Temporal Attention Model for tactile texture recognition.

Nevertheless, these research studies addressed different problems than the one we focus on in this study. Instead, our aim is to identify the most salient contact areas in tactile images obtained from high-resolution optical tactile sensors. This tactile saliency can assist tactile robots in achieving complex tasks in unstructured environments. For example, a robot may need to explore an object's contour \cite{huang2022soft, kappassov2022tactile, lepora2020optimal, lin2022tactilegym2} in clutter using touch. In such scenarios, the identification of tactile saliency is critical since the robot may lack full knowledge of the object and its surrounding stimuli.

\section{Methodology}\label{sec:method}

\subsection{Tactile Saliency Prediction}\label{subsec:gen_fra}
 Given a real grey-scale tactile image $\text{I}=\{\text{I}_{ij}\in[0,1]\mid i\!\!\in\!\!\{1,...,w\},j\!\!\in\!\!\{1,...,h\}\}$ of size $w \times h$ and a target feature type $\text{F} $, we define a \textit{tactile saliency map} ${\text{S}}^\text{F}$ = $\left \{ \text{s}^\text{F}_{ij} \in [0,1] \mid i \in\{1,...,w\},j\in\{1,...,h\}\right\}$ as a matrix of probabilities where $\text{s}^\text{F}_{ij}$ is the probability of pixel $\text{I}_{ij}$ being part of the feature $\text{F}$ (e.g. an edge or surface) in the tactile image~$\text{I}$. We define a mapping $\psi_\text{F}(.): \mathbb{R}^{w\times h}\rightarrow \mathbb{R}^{w\times h}$ as a function that maps a real tactile image $\text{I}$ to a tactile saliency map ${\text{S}}^\text{F}$,
 \begin{align}\label{eq:tacsal}
\psi_\text{F}(\text{I}) := {\text{S}}^\text{F}.
\end{align}
Our aim is to learn a tactile saliency prediction model to approximate $\psi_{\text{F}}$, which can be used to predict a tactile saliency map ${\text{S}}^\text{F}$ representing the probabilities of target feature areas in image $\text{I}$. However, unlike visual saliency datasets, it is challenging for humans to construct a tactile saliency dataset by accurately labelling real tactile images with tactile saliency maps. This is due to the inherent nature of marker-based tactile images, which can present counterintuitive patterns that are difficult for humans to identify and separate the target features from the noise features. Thus, it is impractical to learn a tactile saliency model $\psi_\text{F}$ for $\text{F}$ that directly predicts ${\text{S}}^\text{F}$ from $\text{I}$.

\subsection{Contact Depth Prediction}\label{subsec:con_sal}
While it is challenging to learn $\psi_\text{F}$ to predict $\text{S}^\text{F}$ directly from $\text{I}$, we can alternatively predict $\text{S}^\text{F}$ from a simplified tactile image that only represents contact areas, giving a more straightforward and interpretable representation. Here, we define a simplified tactile image of $\text{I}$ as a \textit{contact depth map} ${\text{C}}$ = $\left \{ \text{c}_{ij} \in [0,1] \mid i \in \{1, \ldots, w\},  j \in \{1, \ldots, h \} \right \}$ where $\text{c}_{ij}$  describes the contact depth level of the tactile skin in pixel $\text{I}_{ij}$. We define a mapping $\phi: \mathbb{R}^{w\times h}\rightarrow \mathbb{R}^{w\times h}$ as a function that maps a real tactile image $\text{I}$ to a contact depth map $\text{C}$,
\vspace{-0cm}
\begin{align}\label{eq:condep_func}
\phi(\text{I}) :=  {\text{C}}.
\end{align}

Our aim is to learn a contact depth prediction model $G_{\text{C}}$ (referred to as \textit{ConDepNet}) to approximate $\phi$ to predict a contact depth map $\text{C}$ representing the contact areas in tactile image~$\text{I}$. Following a general image-to-image translation approach in \cite{isola2017image}, we use pix2pix GAN to learn $G_{\text{C}}$ with an objective in an adversarial training manner,
\begin{align}\label{eq:condep_model}
G_{\text{C}} = G^{\ast }=\text{arg\ }\underset{G}{\text{min}\ }\underset{D}{\text{max}\ }\mathcal{L}_\text{C}(G,D) \text{,}
\end{align}
where $G$ and $D$ are the generator and the discriminator respectively in a GAN setting, with a standard definition
\begin{align}\label{eq:condep_model}
\mathcal{L}_\text{C}(G,D)&=\mathcal{L}_\text{cGAN}(G,D)+\alpha\mathcal{L}_{L_1}(G) \text{,}\\
\mathcal{L}_\text{cGAN}(G,D)&=\mathbb{E}_{\text{I}\in \Gamma,\text{C} \in \Phi} [\text{log}D(\text{I},\text{C})] \nn\\ &\hspace{5em}+\mathbb{E}_{\text{I}\in \Gamma}[\text{log}(1-D(\text{I},G(\text{I}))] \text{,}
\end{align}
where $\Gamma = \left \{ \text{I}_k \mid k \in \{ 1, \ldots, n \} \right \}$ are $n$ real tactile images and $\Phi = \left \{ \text{C}_k \mid k \in \{1, \ldots, n \} \right \}$ are $n$ contact depth maps
, $L_1$  represents the Manhattan norm. Note that, unlike \cite{isola2017image}, we do not consider a random vector as input to $G$, since we need a deterministic contact depth image for each real tactile image. 

To collect a dataset $\mathcal{D}_{\Gamma,\Phi}=\left \{ (\text{I}, \text{C})\mid (\text{I} \in \Gamma, \text{C} \in \Phi  \right \}$ with auto-labelling, we leverage Tactile Gym \cite{church2021tactile}, a simulator for tactile robotics based on rigid-body physics. We use this to generate simulated tactile images rendered as the contact depth map captured by a simulated tactile sensor when contacting stimuli. These contact depth maps serve as a simplified representation of real tactile images, which can be used to label the corresponding contact areas in real tactile images obtained with the same relative contact poses (see Fig.~\ref{fig:method}a). Thus, a set of real tactile images $\Gamma$ labelled with corresponding simulation-generated contact depth maps $\Phi^\Gamma$ is accessible and enables us to learn a ConDepNet without the need for human labelling.

\subsection{Predicting Tactile Saliency from Contact Depth}\label{subsec:ts_cd}
To achieve our primary aim of mapping saliency, we define a mapping $\delta_{\text{F}}: \mathbb{R}^{w\times h}\rightarrow \mathbb{R}^{w\times h}$ as a function that maps a contact depth map $\text{C}$ to a tactile saliency map $\text{S}^\text{F}$ for a given target feature ~$\text{F}$:
\begin{align}\label{eq:con_to_tac}
\delta_\text{F}(\text{C}) := \text{S}^\text{F}.
\end{align}
In other words, we can solve Eq. \ref{eq:tacsal} with a composite function of Eq. \ref{eq:condep_func} and Eq. \ref{eq:con_to_tac}, because a contact depth map $\text{C}$ preserves the contact information in a real tactile image $\text{I}$,
\begin{align}\label{eq:composite}
\delta_\text{F}(\phi(\text{I})) = \delta_\text{F}(\text{C}) = \text{S}^\text{F} = \psi_\text{F}(\text{I}).
\end{align}

Thus our aim reduces to learning a tactile saliency prediction model $G_{\text{S}^\text{F}}$ (referred to as \textit{TacSalNet}) to approximate $\delta_\text{F}$ for a target feature $\text{F}$ that predicts a saliency map $\text{S}^\text{F}$ representing the target areas in $\text{C}$. Similar to ConDepNet, we also apply pix2pix GAN to learn $G_{\text{S}^\text{F}}$, with an objective
\begin{align}\label{eq:condep_model}
G_{\text{S}^\text{F}} = G^{\ast }=\text{arg\ }\underset{G}{\text{min}\ }\underset{D}{\text{max}\ }\mathcal{L}_{\text{S}^\text{F}}(G,D) \text{,}
\end{align}
where $\mathcal{L}_{\text{S}^\text{F}}(G,D)$ is defined like  $\mathcal{L}_{\text{C}}(G,D)$ in Eqs~(4,5) with $\text{S}^\text{F}$ replacing $\text{C}$, drawn from 
$\Psi^\text{F} = \left \{ \text{S}^\text{F} \right \}$ a set of tactile saliency maps for a target feature $\text{F}$. 

A dataset $\mathcal{D}$ consisting of contact depth maps $\Phi$ and their corresponding tactile saliency maps $\Psi^{\text{F}}$ is generated using Tactile Gym as follows (see Fig.~\ref{fig:method}b):\\
\noindent 1) We first collect a set of contact depth maps $\Phi^{\text{F}}=\left \{ \text{C}^\text{F}_k \mid k \in \{ 1, ..,n \} \right \}$ for the target feature type $\text{F}$, and a set of noise contact depth maps $\Phi^{\text{N}}=\left \{ \text{C}^\text{N}_k \mid k \in \{1, ..,m\} \right \}$ for random noise features, where $n$ and $m$ denote the sizes of $\Phi^{\text{F}}$ and $\Phi^{\text{N}}$ respectively.\\
\noindent 2) Then, we generate a set of contact depth maps containing both target and noise features: $\Phi=\left \{ \text{C}_{kl} = \text{C}_{k}^\text{F} + \text{C}_{l}^\text{N} \mid \text{C}_k^\text{F} \in \Phi^\text{F}, \text{C}_l^\text{N} \in \Phi^\text{N} \right \}$, where we overlay $\text{C}_{k}^\text{F}$ with $\text{C}_{l}^\text{N}$ by the pixel-by-pixel sum of pixel values. We label each $\text{C}_{kl} \in \Phi$ with its corresponding $\text{C}_{k}^\text{F} \in \Phi^\text{F}$ to generate a dataset $\mathcal{D}_{\Phi,\text{F}} = \left \{ (\text{C}, \text{C}^\text{F}) \mid \text{C} \in \Phi, \text{C}^\text{F} \in \Phi^{\text{F}}   \right \}$.\\
\noindent 3) Finally, we consider that the saliency map $\text{S}^{\text{F}}$ is a min-max normalized form of $\text{C}^{\text{F}}$, therefore we can obtain $\mathcal{D}_{\Psi,\text{F}} = \left \{ (\text{C}, \text{S}^\text{F}) \mid \text{S}^\text{F} = {\rm Norm}(\text{C}^\text{F}), \text{C} \in \Phi, \text{C}^\text{F} \in \Phi^{\text{F}}  \right \}$ by linearly scaling each $\text{C}^\text{F}$ in $\mathcal{D}_{\Phi,\text{F}}$ to the range of $[0,1]$. Note that while here we only demonstrate a simple and straightforward rescaling method (i.e. Min-Max Normalization) to prove our concept; there are alternative methods like 'standardization' for converting a $\text{C}^{\text{F}}$ to $\text{S}^{\text{F}}$.
\vspace{-0.4em}

\subsection{Tactile Noise Generator}\label{subsec:tng}
We could use a simulated tactile sensor to interact with noise stimuli in Tactile Gym to collect a set of noise contact depth maps $\Phi^{\text{N}}$ for learning a TacSalNet. However, this is inefficient as various stimuli CAD models are required. Also, the noise patterns are unlike those we see in contact depth maps generated from a ConDepNet in real-world experiments. Therefore, we propose a generative model $\tau$ to generate random noise features, which we call Tactile Noise Generator (TacNGen), as shown in Fig.~\ref{fig:method}c. Specifically, We use a conventional structure of Variational AutoEncoder (VAE) to learn an encoder parameterised by $\theta$: $f_{\theta}(\text{C}) = (\mu_{\theta , \text{C}}, \Sigma_{\theta , \text{C}})$ where $\mu_{\theta , \text{C}} \in \mathbb{R}^K$ and $\Sigma_{\theta , \text{C}} = \text{diag}(\sigma_1^2, .., \sigma_K^2)$, and a decoder parameterised by $\lambda$: $f_{\lambda}(z)=\hat{\text{C}}$ to reconstruct a contact depth map $\hat{\text{C}}$ from a $K$-dim latent vector $z \sim \mathcal{N}(\mu_{\theta , \text{C}}, \Sigma_{\theta , \text{C}})$. Thus, we construct a loss function $\mathcal{L}_\text{TN}$:
\begin{align}\label{eq:TN_loss}
\mathcal{L}_\text{TN} = \mathbb{E}_{\text{C}\in \Phi^\text{N}}[\alpha\text{KL}[p_{\theta}(z \mid \text{C})\parallel q(z)]-\mathbb{E}_{p_{\theta}(z \mid \text{C})}\text{log}p_{\lambda}(\text{C} \mid z)]\text{.}
\end{align}
The regularization term $\text{KL}[p_{\theta}(z \mid \text{C})\parallel q(z)]$ is the Kullback–Leibler divergence between $p_{\theta}(z \mid \text{C}) = \mathcal{N}(\mu_{\theta , \text{C}}, \Sigma_{\theta , \text{C}})$ and $q(z) = \mathcal{N}(0,\mathbb{I})$ where $\mathbb{I}$ is an identity matrix. The second term $\mathbb{E}_{p_{\theta}(z \mid \text{C})}\text{log}p_{\lambda}(\text{C} \mid z)$ is the reconstruction log-probability; we control the trade-off between these two terms with a hyperparameter $\alpha$. After minimizing $\mathcal{L}_\text{TN}$ with dataset $\Phi^\text{N}$ by simultaneously optimizing with respect to $\theta$ and $\lambda$, we can then obtain a TacNGen $\tau$ to generate a random noise contact depth map $\hat{\text{C}^\text{N}}$ with
\begin{align}\label{eq:TN_loss}
\tau(z)=f_{\lambda^{\ast }}(z) = \hat{\text{C}^\text{N}}\text{ , } 
\end{align}
where $ z \sim \mathcal{N}(\mu, \Sigma)$ and $\lambda^{\ast}=\text{argmin}_{\theta,\lambda}\mathcal{L}_\text{TN}$.

\section{Experiments and Results} \label{sec:expe}
\subsection{Preliminary} \label{subsec:preliminary}
\subsubsection{Hardware}  
For all experiments in this paper, we use a desktop robot (Dobot MG400), which is a relatively low-cost, high-accuracy robot well-suited for tactile robot control tasks \cite{lin2022tactilegym2,lepora2022digitac}. We attach a low-cost, high-resolution optical tactile sensor called the TacTip~\cite{lepora2021soft} as an end-effector. This tactile sensor features an internal array of markers on biomimetic protruding pins inside the soft tactile skin to capture marker-based movement that amplifies the skin deformation induced by physical contact against external stimuli. 

\subsubsection{Dataset and Networks Training} \label{subsubsec:data_collection}

To streamline our analysis, in this paper we limit our focus to an edge feature as the target feature. To this end, we first collect 7000 tactile images of edge-feature data by conducting labelled random contacts between the TacTip and a square stimulus with flat edges (Fig.~\ref{fig:pose_acc}a, top image). These contacts are made within ranges of position $y\in[-6,6]$\,mm, depth $z\in[3,6]$\,mm, and rotation $R_z \in [-180^{\circ},180^{\circ}]$ both in the real world and in simulation. Then we develop a TacNGen in simulation for creating various noisy contact depth maps. Specifically, we train a VAE with conv layers to reconstruct contact depth maps of simulated cone-shaped distractors that vary in radii and angles (see Fig. \ref{fig:method}c). Further VAE training hyperparameters can be found in the supplementary files. In practice, we found that generating the noise tactile images from TacNGen and then applying conventional image augmentation to them in every epoch of training the TacSalNet resulted in similar performance compared to regenerating the noise tactile images every epoch. With these datasets and the proposed methods (Secs \ref{subsec:con_sal},\ref{subsec:ts_cd}), we can learn a ConDepNet and a TacSalNet using pix2pixGAN \cite{isola2017image} with hyperparameters specified in \cite{church2021tactile}. It is important to note that the real tactile images are used only for the ConDepNet, while the TacSalNet and the TacNGen are trained solely in simulation.

\subsection{TacSalNets with Different Tactile Noise Generation}\label{subsec:pose_pred_exp}

\begin{figure}
  \centering
    \includegraphics[width=.9\linewidth]{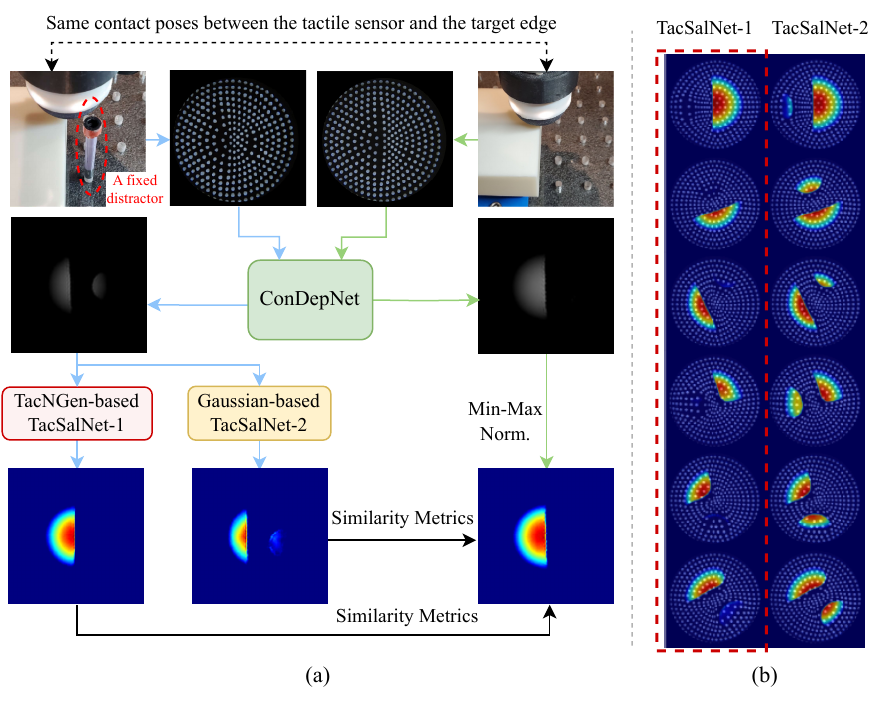}
    \vspace{-1em}
    \caption{(a) Illustration of the evaluation process for TacSalNets trained with TacNGen-generated noise (TacSalNet-1) and Gaussian distribution noise (TacSalNet-2). Blue arrows indicate the pipeline for predicted saliency map generations, while green arrows show the pipeline for target saliency map generation. (b) Tactile saliency maps predicted from TacNGen-based (left) and Gaussian-based (right) TacSalNets respectively using the same input tactile images (overlay) in each row.}
  \label{fig:tacngen_abl}
  \vspace{-1em}
\end{figure}

 First, we conduct an ablation study to compare the performance of our tactile saliency network (TacSalNet) when the noise is generated using our proposed noise generation model (TacNGen) versus a 2-dim multivariate Gaussian distribution. For conciseness, we will refer to the TaSalNet trained with noise generated from TacNGen as \textbf{TacSalNet-1}, and the one trained with Gaussian noise as \textbf{TacSalNet-2}. The evaluation process, depicted in Fig.~\ref{fig:tacngen_abl}a, uses the same pose ranges as those used during data collection (see Sec.~\ref{subsubsec:data_collection}). We evaluate the performance of both TacSalNets with four commonly-used metrics in visual saliency research: AUC-Judd, similarity measure (SIM), Pearson's linear coefficient (CC), and normalized scanpath saliency (NSS), and the testing results are reported in Table~\ref{table:tacngen_abl}. The results demonstrate that the TacSalNet-1 outperforms the TacSalNet-2 in three metrics and one the same, indicating the effectiveness of our TacNGen in improving the performance of tactile saliency prediction compared to domain-agnostic stochastic noise generation. Additionally, we found that even though they are only trained with straight-edge features, the TacSalNet-1 can generalize well to unseen corner-edge features while the TacSalNet-2 tends to preserve the noise features (see the last 3 rows in Fig.~\ref{fig:tacngen_abl}b). Thus, we only consider the TacSalNet-1 (TacNGen-based) for the evaluation of our whole framework in the contact pose prediction task and the edge-following task.

\begin{table}[]
\centering
\caption{Real-world validation with similarity metrics for TacSalNet-1 and TacSalNet-2 over 1000 samples. Bold numbers denote the best results.}
\begin{tabular}{c|cccc}
\hline
            & AUC-J $\uparrow $          & SIM $\uparrow $            & CC $\uparrow $             & NSS $\uparrow $            \\ \hline
TacSalNet-1 & \textbf{0.995} & \textbf{0.984} & \textbf{0.957} & \textbf{4.629} \\
TacSalNet-2 & \textbf{0.995} & 0.972          & 0.936          & 4.288          \\ \hline
\end{tabular}
\label{table:tacngen_abl}
\vspace{-0.5em}
\end{table}

\subsection{Tactile Saliency Prediction for Contact Pose Estimation}\label{subsec:pose_pred_exp}

\begin{figure*}
  \centering
    \includegraphics[width=.8\linewidth]{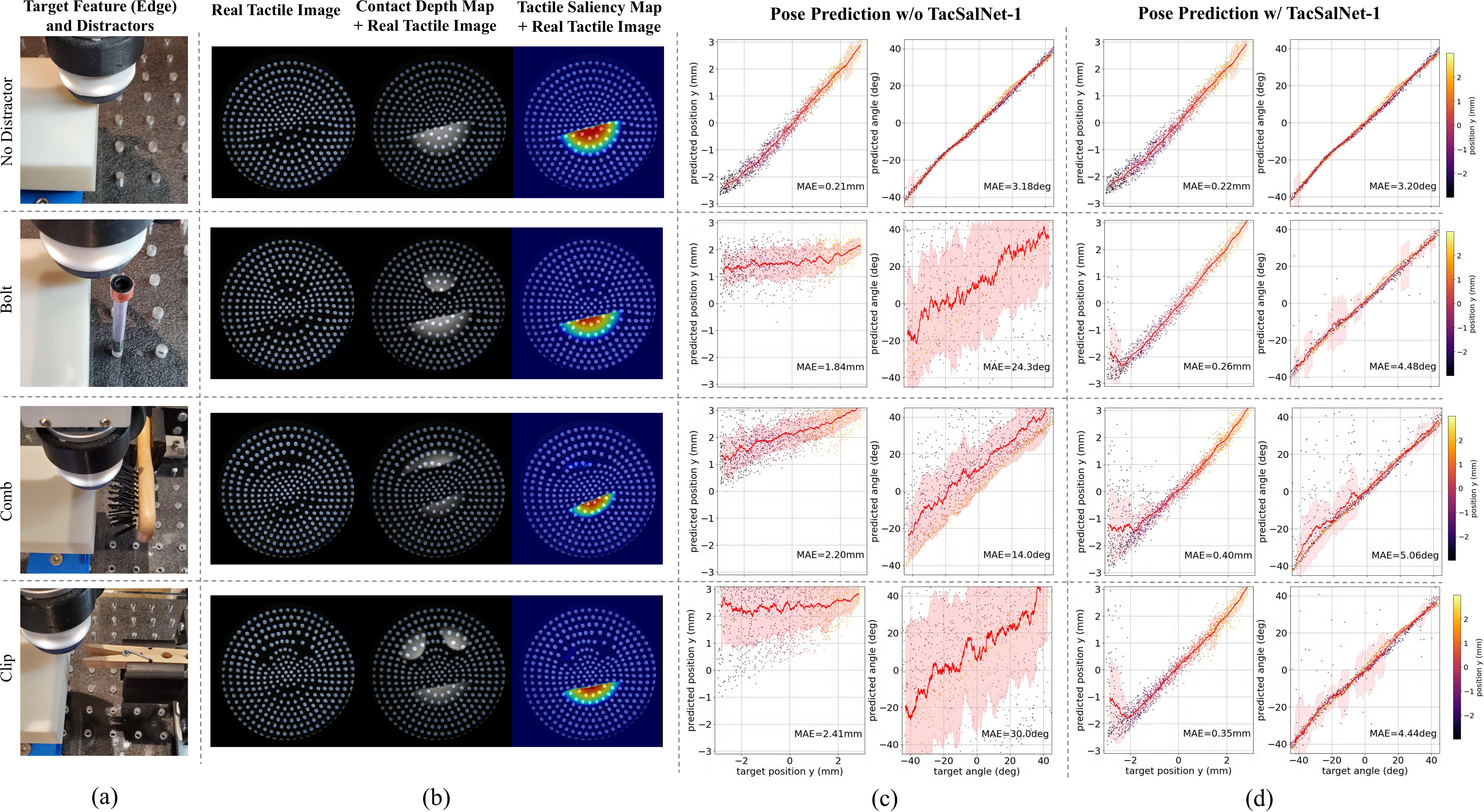}
    \vspace{-0.5em}
    \caption{Real-world evaluation of tactile saliency prediction based on the tactile PoseNet prediction accuracy on the target feature (edge) distracted by various everyday objects: (a) the distraction setup with everyday objects of various shapes; (b) real tactile images of the local contact paired with related contact depth map and tactile saliency map; (c) tactile PoseNet prediction accuracy on 2D edge pose without and (d) with tactile saliency prediction.}
  \label{fig:pose_acc}
\end{figure*}

In our second experiment, we evaluate the performance of tactile saliency prediction in improving contact pose prediction accuracy in the presence of distractors. To achieve this, we apply the TacSalNet-1 to a tactile PoseNet\footnote{Here we use a sim-to-real version of the PoseNet: \url{https://github.com/ac-93/tactile_gym_servo_control}.}, which is a Convolutional Neural Network that predicts pose from tactile images~\cite{lepora2020optimal}, and investigate its impact on predicting the target edge 2D pose (position $y$ and orientation $R_z$) from a tactile image when the TacTip statically contacts the target edge and distractors (see Fig.~\ref{fig:pose_acc}a). In particular, we consider five everyday objects as distractors varied in shapes, including a bolt, a comb with soft tines, a wooden clip, a toy figure, and a spoon (the results of the last two objects are shown in the supplementary materials). Each distractor is fixed next to the target edge, with distances ranging between $[7,14]$\,mm. The contact poses are sampled from position $x \in[-10,10]$\,mm,  $y \in[-3,3]$\,mm, $z\in[3,6]$\,mm, and rotation $R_z \in [-45^{\circ},45^{\circ}]$.

The results show that the presence of the distractors significantly impairs the standard PoseNet performance, resulting in mean absolute errors (MAEs) of $1.84$-$2.41$\,mm and $14.0^{\circ}$-$30.0^{\circ}$ for predicting position $y$ and orientation $Rz$ respectively (Fig.~\ref{fig:pose_acc}c). However, when the TacSalNet-1 is applied, the PoseNet can maintain its accuracy with position $y$ MAE of $0.26$-$0.40$\,mm and orientation $R_z$ MAE of $4.44^{\circ}$-$5.06^{\circ}$ (Fig.~\ref{fig:pose_acc}d, note that we calibrated the unavoidable prediction drift caused by the distractors). Importantly, tactile saliency prediction does not degrade the PoseNet performance, as evident from the marginal difference in MAEs between the results with and without TacSalNet-1 is minor (the first row in Fig.~\ref{fig:pose_acc}).

To demonstrate the generalizability of our tactile saliency prediction method on various real-world noise features, we randomly selected real tactile images with paired contact depth maps and tactile saliency maps induced by static contact with different distractors (Fig.~\ref{fig:pose_acc}b). In the last pair of tactile images of Fig.~\ref{fig:pose_acc}b, we see the TacSalNet-1 accurately predicts the target edge shape even in the presence of distracting contacts that were unseen during training. 

We observe that the performance of the PoseNet with tactile saliency prediction drops in all cases when the TacTip position $y$ is around $-2.5$\,mm. We attribute this to the close proximity of the TacTip to the distractors, and accordingly the TacSalNet-1 tends to instead predict the distractor as part of the target feature, causing a decrease in PoseNet accuracy.

\subsection{Tactile Saliency Prediction in the Edge Following Task}\label{subsec:edge_follow_exp}

\begin{figure}
  \centering
    \includegraphics[width=1\linewidth]{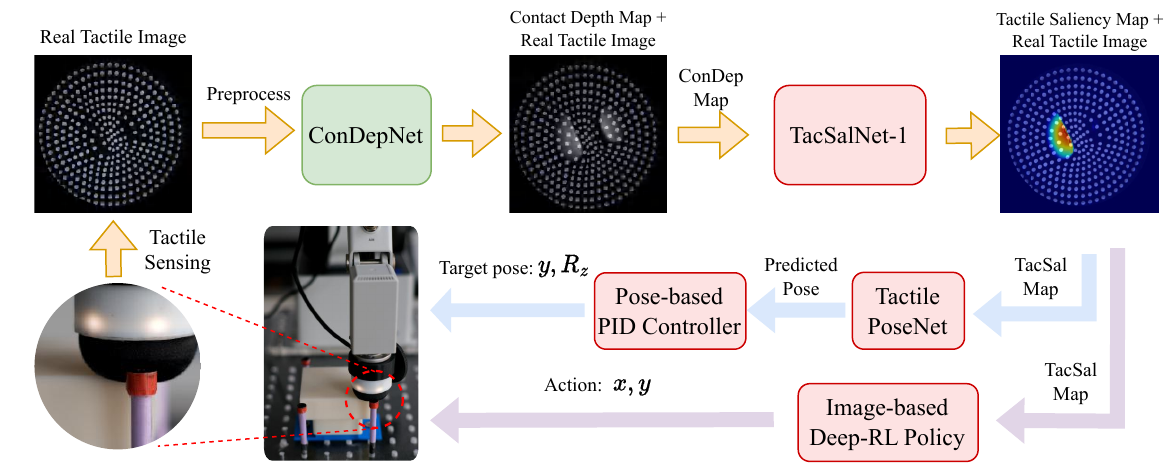}
    \vspace{-0.5em}
    \caption{The diagram of two types of sim-to-real tactile control methods augmented with tactile saliency prediction for the edge-following task with unexpected distractors. The orange arrows denote the pipeline of tactile image processing, the blue and purple arrows denote the pipelines of the pose-based PID control and image-based deep RL policy respectively. The network in green is trained with real-world and simulated data, while the ones in red are obtained solely in simulation and applied to the real world without fine-tuning. } 
  \label{fig:con_fra}
  \vspace{-0.2em}
\end{figure}

\begin{table}[]
\vspace{-0.5em}
\caption{MAEs of the trajectories from the ground truth for the edge-following task with distractors using pose-based PID control and Image-based deep RL policy without and with TacSalNet-1 (short for TSN-1 in this table).}
\vspace{-0.5em}
\centering
\begin{tabular}{c|cc|cc}
\hline
\multirow{2}{*}{Objects} & \multicolumn{2}{c|}{\textbf{Pose-based PID}} & \multicolumn{2}{c}{\textbf{Image-based deep RL}} \\
                         & w/o TSN-1           & w/ TSN-1           & w/o TSN-1          & w/ TSN-1           \\ \hline
Square                   & Fail              & 0.77mm          & Fail             & 1.04mm          \\
Foil                     & Fail              & 0.53mm          & Fail             & 0.94mm          \\
Flower                   & Fail              & 0.76mm          & Fail             & 1.03mm          \\
Volute                   & Fail              & 0.91mm          & Fail             & 1.21mm          \\ \hline
\end{tabular}
\label{table:edge_follow_acc}
\vspace{-1em}
\end{table}

\begin{figure*}
  \centering
    \includegraphics[width=0.9\linewidth]{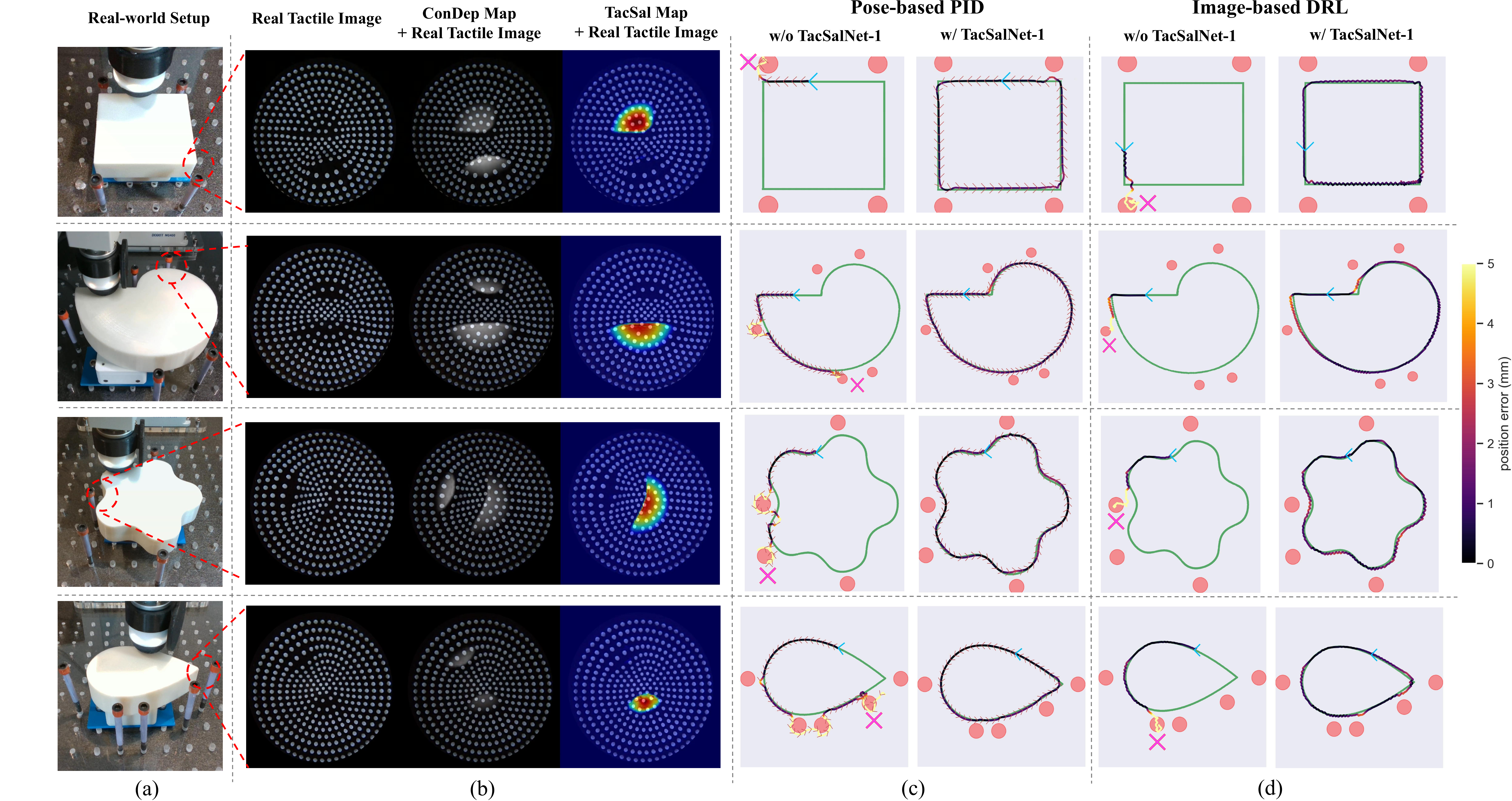}
    \vspace{-0.5em}
    \caption{Evaluation of the proposed tactile-saliency-based control framework in the real-world 2D edge-following task: (a) the target objects are surrounded with distractors; (b) the TacSalNet-1 can accurately predict the target-feature shapes that are unseen during training; edge-following performance for both (c) pose-based PID control and (d) image-based deep-RL control: the green contour shows the ground truth for each target object; errors of the traced contour from the ground truth are colour-coded (side colour bar); the purple crosses indicate the distractors that the TacTip remained stuck within 50 steps; the blue arrow denotes the starting pose; and the red arrow denotes the orientation of the TacTip with the pose-based PID controller (the image-based deep-RL policy only outputs x and y actions).}
  \label{fig:edge_follow}
  \vspace{-0.85em}
\end{figure*}

In our third experiment, we investigate the performance of the tactile saliency prediction in improving the robustness of different tactile control methods during an edge-following task that involves distractors. Specifically, the robot is tasked to control its tactile sensor to follow along the edges of four target objects chosen to have distinct edge features (a square, volute, flower and foil). These target objects are surrounded by at least four distractors fixed next to their edges with distances between $[7,12]$\,mm (Fig.~\ref{fig:edge_follow}a). In this task, we focus on the effect of the distractor as the edge changes in curvature, considering just one type (the bolt) given the effectiveness of the TacSalNet-1 for all distractors considered in Sec.~\ref{subsec:pose_pred_exp}.   

We consider two distinct state-of-the-art tactile control methods: 1) tactile pose-based PID control \cite{lepora2021pose}, and 2) tactile image-based deep reinforcement learning \cite{church2021tactile}. Note that both the pose-based PID controller and the image-based deep-RL policy are learned solely in simulation with the contact depth map as input, and apply to the real world without fine-tuning. To evaluate whether our proposed method can improve the robustness of these control methods in the presence of distractors, we augment them with TacSalNet-1 (pipeline in~Fig.~\ref{fig:con_fra}). 

The experimental results (Fig.~\ref{fig:edge_follow}) indicate that the use of either tactile control methods alone in the presence of distractors is insufficient to successfully complete the task, as the robot becomes distracted by bolts and fails away from the edge of the target objects. Thus neither control method alone is robust to unexpected distractions. However, when the TacSalNet-1 is used to predict tactile saliency to augment the tactile input to both control methods, the robot successfully achieves the task by completing a circuit of the object. 

The accuracies of the pose-based PID controller and the image-based deep-RL policy with tactile saliency prediction from 80 real-world tests (10 for each object with each method combination) are 0.53-0.91 \,mm and 0.94-1.21\,mm respectively (Table \ref{table:edge_follow_acc}), compared to an overall distance travelled of 300-520\,mm. Overall, the tactile saliency prediction significantly improves the robustness of sim-to-real tactile control.  

Moreover, we observed that the sim-to-real tactile control methods with saliency prediction performed consistently well on all objects, despite variations in their edge features, such as sharp and right-angle corners, concave and convex edges with noise features (Fig.~\ref{fig:edge_follow}a). This finding indicates the generalizability of our saliency prediction framework, which is a key advantage for tactile robotic systems that operate in diverse and unstructured environments. A broader range of distracting objects was also considered with this edge-following task, and the method found to be similarly robust to distractors; the videos for these experiments are available in the supplementary materials.



\section{DISCUSSION AND FUTURE WORK} \label{sec:discussion}

In this work, we introduce the concept of \textit{tactile saliency} inspired by visual saliency, to describe the target features from real tactile images captured by optical tactile sensors, for improving the accuracy and robustness of tactile control in the presence of distractors. To develop a generalizable tactile saliency prediction network, we propose a 3-stage approach. First, we employ a ConDepNet to generate contact depth maps, which can extract the global contact information that contains target and noise features in real tactile images. Second, we introduce a TacSalNet to predict a tactile saliency map that accurately describes the target areas within an input contact depth map. Finally, we develop a TacNGen, which generates various noise features to train the TacSalNet, thereby improving its generalizability to unseen contact depth maps. Notably, we only train the ConDepNet using real-world data, while the TacSalNet and TacNGen are trained solely in simulation.

We validated our approach in the real world by demonstrating the efficacy of tactile saliency prediction alongside previous works in: 1) improving tactile pose prediction accuracy \cite{lepora2020optimal} in the presence of different distractors (Sec. \ref{subsec:pose_pred_exp}); 2) enhancing the robustness of two different sim-to-real tactile control methods, namely pose-based PID control \cite{lepora2021pose} and image-based deep reinforcement learning \cite{church2021tactile}, during an edge-following task that involves distractors (Sec. \ref{subsec:edge_follow_exp}).

Our proposed approach offers several key advantages. Firstly, it eliminates the need for labour-intensive human labelling
, resulting in a more efficient and practical data collection process. Secondly, with the high fidelity of generated tactile noise, it enables accurate prediction of target features from contact depth maps in the presence of unseen noise, enhancing the accuracy of the contact pose prediction and the robustness of tactile control in unstructured environments. Thirdly, the TacSalNet can be seen as a simple plug-in module that can extend readily for various types of sim-to-real tactile control methods, such as the ones we considered here \cite{lepora2021pose, church2021tactile}. 


Our approach was tested with a marker-based optical tactile sensor (the TacTip) and relies on a depth-based tactile simulator (Tactile Gym). Previous work \cite{lin2022tactilegym2} has demonstrated that different tactile sensors (DIGIT \cite{lambeta2020digit} and DigiTac \cite{lepora2022digitac}) can be easily incorporated into tactile gym for sim-to-real tactile policy learning. It will be interesting in future work to explore how our approach applies to other types of tactile sensors.


Overall, we expect the tactile saliency approach developed here can be applied more widely to various other exploration and manipulation tasks involving optical tactile sensors that are adversely affected by unexpected tactile distractors.

\bibliographystyle{IEEEtran}
\bibliography{IEEEabrv,manual}

\end{document}